\documentclass{article}

\usepackage{iclr2026_conference,times}

\usepackage{amsmath,amsfonts,bm}

\def\eqref#1{equation~\ref{#1}}

\def\1{\bm{1}}

\def\F2{{$\mathbb{F}_2$}}

\DeclareMathAlphabet{\mathsfit}{\encodingdefault}{\sfdefault}{m}{sl}
\SetMathAlphabet{\mathsfit}{bold}{\encodingdefault}{\sfdefault}{bx}{n}

\usepackage{amsmath,amssymb,amsthm}
\usepackage{graphicx}
\usepackage{import}
\usepackage{hyperref}
\usepackage{listings}
\usepackage{booktabs}

\usepackage[utf8]{inputenc}
\usepackage[T1]{fontenc}
\usepackage{textcomp}
\usepackage{yfonts}
\newcommand{\longs}{{\textfrak{s}}}

\newcommand{\glitchtoken}{%
  \texttt{\_\raisebox{-0.5ex}{\includegraphics[height=2.5ex]{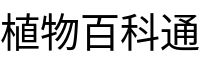}}}%
}

\title{Check Your LLM's Secret Dictionary!\\ Five Lines of Code Reveal What Your LLM Learned (Including What It Shouldn't Have)}

\author{Hisashi Miyashita \\
Mgnite Inc.\\
\texttt{himi@mgnite.com}
}

\iclrfinalcopy %
\begin{document}

\maketitle

\begin{abstract}
We show that singular value decomposition of the \texttt{lm\_head} weight matrix of a transformer-based large language model---requiring only five lines of PyTorch and no model inference---reveals interpretable semantic subspaces directly from the model weights.
Each left singular vector identifies the vocabulary tokens most readily selected when the hidden state aligns with the corresponding singular direction; inspecting these clusters exposes the model's training data composition and curation philosophy.

Analysing GPT-OSS-120B, Gemma-2-2B, and Qwen2.5-1.5B, we find that singular value spectra and vocabulary cluster structures differ systematically across models: GPT exhibits a graduated hierarchy of functionally differentiated subspaces; Gemma is dominated by pre-nineteenth-century English orthography, forming a stepwise clustering structure that may contribute to high output controllability; and Qwen exhibits broad multilingual coverage alongside subspaces whose vocabulary the authors have determined to be ethically inappropriate for direct publication.

Base--instruct comparison reveals that ethically concerning subspaces originate in pretraining and are not removed by post-training alignment.
We introduce the Vocabulary Cluster Score (VCS) to quantify subspace coherence, and the Weighted Projection Score (WPS) as a static glitch token detector; applying WPS to GPT-OSS-120B recovers \glitchtoken{} (ID~137606), a well-known glitch token widely reported in the CJK language community, without any model inference.
We propose a taxonomy of root causes for problematic vocabulary content and call for \texttt{lm\_head} SVD analysis to be adopted as a standard pre-release safety auditing step.
Our findings further suggest directions toward SVD-guided tokenizer optimisation and more controllable LLM design.
\end{abstract}

\section{Introduction}

The output projection layer of a transformer-based large language model---commonly referred to as \texttt{lm\_head}---is a weight matrix $W \in \mathbb{R}^{V \times d}$ that maps hidden states to logits over the vocabulary of size $V$. It is the final gate through which every token the model produces must pass, yet it is rarely analysed in isolation. In this paper, we ask a simple question: \emph{what does the \texttt{lm\_head} weight matrix itself reveal about a model, before any inference is run?}

The answer turns out to be surprisingly rich. Applying singular value decomposition (SVD) to $W$ and inspecting the top-scoring tokens of each left singular vector requires only five lines of PyTorch. Yet the resulting vocabulary clusters are interpretable, systematic, and---in at least one case---alarming. The singular vectors partition the vocabulary into groups that reflect the model's training data composition, curation philosophy, and, as we show, the presence of content that the authors have determined to be ethically inappropriate for direct publication.

\paragraph{Key findings.}
We analyse three publicly available models---GPT-OSS-120B, Gemma-2-2B, and Qwen2.5-1.5B---and report the following.

\begin{enumerate}

\item \textbf{lm\_head SVD reveals what a model is most likely to say, statically.}
Each left singular vector $U[:,i]$ identifies the vocabulary tokens that are most readily selected when the hidden state aligns with the $i$-th singular direction. Because RMSNorm normalises the hidden state onto the unit sphere before the \texttt{lm\_head} projection, only the directional structure of the hidden state matters; the singular directions are thus the fundamental axes of vocabulary selection. This analysis requires no model inference and no input data.

\item \textbf{The structure of these axes varies systematically across models.}
GPT-OSS-120B exhibits a graduated hierarchy of functionally differentiated subspaces---syntactic structure, attribute taxonomy, formal specification vocabulary, software engineering terminology---with a gentle decay in singular values. Gemma-2-2B is dominated by historical English orthography (pre-nineteenth-century long-s typography), forming a stepwise clustering structure in the singular value spectrum that may contribute to high output controllability. Qwen2.5-1.5B exhibits script-level clustering alongside geometrically dispersed subspaces spanning unrelated languages and domains. These differences directly reflect each model's training data composition and curation philosophy.

\item \textbf{Post-training alignment (RLHF) does not resolve pretraining-level issues.}
Comparing base and instruction-tuned variants of Qwen2.5-1.5B and Gemma-2-2B, we find that the \texttt{lm\_head} structure of Gemma is virtually unchanged by RLHF, while Qwen undergoes substantial remapping of symbol clusters to multilingual lexical clusters. Critically, ethically concerning vocabulary subspaces---where present---persist across both base and instruct variants with comparable structure, demonstrating that their root cause lies in pretraining data rather than post-training procedure.

\item \textbf{lm\_head SVD analysis is a viable safety auditing tool.}
The method surfaces ethically concerning vocabulary without adversarial prompting, red-teaming, or inference. We propose a taxonomy of root causes for such observations and call for \texttt{lm\_head} SVD analysis to be adopted as a standard pre-release auditing step.

\end{enumerate}

\paragraph{Additional contributions.}
Beyond these primary findings, our analysis yields two further contributions. First, the same SVD decomposition gives rise to the Weighted Projection Score (WPS), defined as $\text{WPS}(v) = \sum_k S[k]\cdot|U[v,k]|$ for each token $v$. Tokens with low WPS are geometrically marginalised in the \texttt{lm\_head} weight space and are candidates for glitch token status~\citep{rumbelow2023solidgoldmagikarp}. Applying WPS to GPT-OSS-120B recovers \glitchtoken{} (ID~137606)---a glitch token widely known in the CJK language community\footnote{See \url{https://note.com/xcloche/n/n55938e706986} for a detailed account in Japanese.}---without any model inference (Section~\ref{sec:glitch}). Second, the Vocabulary Cluster Score (VCS) introduced in this paper provides a quantitative signal that may guide tokenizer optimisation: iteratively retaining high-VCS tokens and pruning low-VCS ones could yield smaller, more geometrically coherent vocabularies. The structural independence of Algebraic Ontology Projection (AOP)~\citep{aop2025} from \texttt{lm\_head} eigenvector pathologies further suggests that AOP-based prompt optimisation may serve as a principled mitigation strategy for the output instability we identify.

\paragraph{Reproducibility.}
The complete analysis requires only five lines of PyTorch, uses only the \texttt{lm\_head} weight matrix and the model tokenizer, and runs in minutes on a single GPU. We release all analysis code to facilitate adoption.

\section{Method}

\subsection{SVD Analysis of \texttt{lm\_head}}

Let $W \in \mathbb{R}^{V \times d}$ denote the \texttt{lm\_head} weight matrix of a transformer-based LLM, where $V$ is the vocabulary size and $d$ is the hidden dimension. We compute the economy SVD:

\begin{equation}
    W = U S V^{\top}
\end{equation}

where $U \in \mathbb{R}^{V \times d}$ contains the left singular vectors, $S \in \mathbb{R}^{d \times d}$ is a diagonal matrix of singular values in descending order, and $V^{\top} \in \mathbb{R}^{d \times d}$ contains the right singular vectors.

Each column $U[:, i]$ assigns a scalar score to every token in the vocabulary. The geometric interpretation is as follows: when the hidden state $h \in \mathbb{R}^d$ (after RMSNorm) is projected along the $i$-th right singular direction $V[:, i]$, the resulting logit contribution is amplified by $S[i]$ and distributed across the vocabulary according to $U[:, i]$. Tokens with the highest scores in $U[:, i]$ are therefore the most likely to be selected when the hidden state aligns with that singular direction. RMSNorm normalizes each element of $h$ onto the unit sphere prior to re-scaling by learned weights~\citep{zhang2019root}, so the magnitude of $h$ is discarded and only its direction determines vocabulary selection. The singular directions of $W$ are thus the fundamental axes of this selection process.

We extract the top-$k$ tokens for each of the first $n$ singular vectors and decode them using the model's tokenizer. This requires no model inference and no input data: only the weight matrix $W$ and the tokenizer are needed. The implementation reduces to the following:

\begin{lstlisting}[language=Python]
U, S, Vh = torch.linalg.svd(
    lm_head_weight.float(), full_matrices=False)
for i in range(n_vectors):
    top_indices = U[:, i].topk(k).indices
    tokens = tokenizer.decode(top_indices)
\end{lstlisting}

\subsection{Vocabulary Cluster Score (VCS)}
\label{sec:vcs}

To quantify the geometric coherence of the vocabulary cluster associated with each singular vector, we introduce the Vocabulary Cluster Score (VCS). This metric adapts the mean pairwise cosine similarity used as a coherence measure in topic modelling~\citep{roder2015topiccoherence} and word embedding evaluation~\citep{schnabel2015evaluation} to the specific geometry of the \texttt{lm\_head} weight space. We note that while cosine similarity is an imperfect measure of semantic similarity in general embedding spaces~\citep{schnabel2015evaluation}, it is particularly well-suited here: the \texttt{lm\_head} computes token logits as the dot product $h \cdot W_v = \|h\|\|W_v\|\cos\theta$, so cosine similarity between two \texttt{lm\_head} row vectors directly reflects the degree to which the model treats the corresponding tokens as contextually interchangeable---a property that is precisely what VCS is intended to capture. For the $i$-th singular vector, VCS is defined as the mean pairwise cosine similarity among the \texttt{lm\_head} row vectors of the top-$k$ ranked tokens:

\begin{equation}
    \text{VCS}(i) = \frac{2}{k(k-1)} \sum_{a < b} \frac{w_a \cdot w_b}{\|w_a\| \|w_b\|}
\end{equation}

where $w_j = W[t_j, :]$ is the \texttt{lm\_head} row vector of the $j$-th top-ranked token $t_j$.

A high VCS indicates that the top-ranked tokens occupy a compact region of the embedding space. This implies that small perturbations of the hidden state along that singular direction will consistently activate tokens within the same geometric neighborhood, resulting in stable and predictable output. A low VCS indicates a geometrically dispersed cluster: small perturbations may activate unrelated tokens, a condition directly associated with output instability and reduced auditability.

We note that VCS measures geometric coherence in the \texttt{lm\_head} weight space, which may reflect semantic similarity, script-level similarity, or functional similarity. Distinguishing between these cases requires qualitative inspection of the decoded tokens, and we perform this inspection for all models in Section~\ref{sec:results}.

\subsection{Base--Instruct Comparison}
\label{sec:base_instruct}

To investigate the effect of post-training alignment on \texttt{lm\_head} structure, we apply the same SVD analysis to both the base and instruction-tuned variants of Qwen2.5-1.5B and Gemma-2-2B. For GPT-OSS-120B, only the instruction-tuned variant is publicly available; base--instruct comparison was therefore not possible for this model.

We compare (i) the decoded token clusters for corresponding singular vectors, and (ii) the VCS values, between base and instruct variants. Changes in token clusters reveal vocabulary remapping induced by post-training alignment; changes in VCS values reveal structural reorganization of the geometric relationships among vocabulary embeddings.

\subsection{Models and Experimental Setup}

We analyze the following models:

\begin{table}[h]
\centering
\caption{Models analyzed in this study.}
\label{tab:models}
\begin{tabular}{llrr}
\toprule
Model & Variant(s) & $V$ & $d$ \\
\midrule
GPT-OSS-120B    & Instruct only  & 201,088 & 2,880 \\
Gemma-2-2B      & Base + Instruct & 256,000 & 2,304 \\
Qwen2.5-1.5B    & Base + Instruct & 151,936 & 1,536 \\
\bottomrule
\end{tabular}
\end{table}

All SVD computations use \texttt{torch.linalg.svd} with \texttt{full\_matrices=False} on float32 weight matrices. We report results for the top $n=30$ singular vectors with $k=20$ top tokens per vector. Experiments were conducted on a DGX Spark system with an NVIDIA GB10 GPU (128\,GB unified memory).

\section{Results}
\label{sec:results}

\subsection{Overview of Eigenvector Structure}

Table~\ref{tab:vcs_summary} summarises the Vocabulary Cluster Score (VCS) statistics for the top 30 singular vectors of each model. Despite differences in architecture and scale, all three models exhibit a consistent pattern: the leading singular vectors attain markedly higher VCS than the tail, indicating that the most influential directions in vocabulary space are also the most geometrically coherent.

\begin{table}[h]
\centering
\caption{VCS statistics over the top 30 singular vectors (instruct variants). $i^{*}$ denotes the index of the singular vector attaining the maximum VCS.}
\label{tab:vcs_summary}
\begin{tabular}{lrrrl}
\toprule
Model & Mean VCS & Max VCS & $i^{*}$ & Dominant cluster type \\
\midrule
GPT-OSS-120B   & 0.14 & 0.30 & 0  & Punctuation / structural tokens \\
Gemma-2-2B     & 0.39 & 0.88 & 4  & Historical English orthography \\
Qwen2.5-1.5B   & 0.25 & 0.64 & 7  & Hangul script characters \\
\bottomrule
\end{tabular}
\end{table}

The dominant cluster type for the highest-VCS eigenvector differs qualitatively across models, reflecting systematic differences in training data composition discussed in Section~\ref{sec:safety}.

\subsection{GPT-OSS-120B}
\label{sec:gpt}

The leading singular vector ($i=0$, VCS$=0.30$) is dominated by punctuation and structural tokens common in English prose and code (\texttt{,}, \texttt{(}, \texttt{in}, \texttt{and}, \texttt{the}, \texttt{a}, \texttt{to}, \texttt{for}). This suggests that the most influential axis of vocabulary selection in GPT-OSS-120B encodes syntactic structure rather than lexical semantics.

The subsequent singular vectors reveal a rich taxonomy of functional subspaces:

\begin{itemize}
    \item $U[:,2]$ (VCS$=0.10$): Attribute and category nouns (\textit{Colours}, \textit{Ages}, \textit{Batteries}, \textit{Bench}, \textit{Arms}), directly corresponding to the \textit{is-a}/\textit{has-a} relational structure exploited by Algebraic Ontology Projection~\citep{aop2025}.
    \item $U[:,7]$ (VCS$=0.24$): Formal specification vocabulary (\texttt{INTERNAL}, \texttt{VALID}, \texttt{RESULT}, \texttt{REQUIRED}, \texttt{CURRENT}, \texttt{OPTION}), consistent with systems engineering and audit terminology.
    \item $U[:,8]$ (VCS$=0.19$): Software engineering vocabulary (\textit{Instances}, \textit{Arguments}, \textit{Dependencies}, \textit{Errors}, \textit{Constraints}).
    \item $U[:,5]$, $U[:,6]$: Simplified and traditional Chinese characters, indicating substantial multilingual training data.
    \item $U[:,25]$ (VCS$=0.28$): Purely numeric tokens, forming a geometrically coherent but semantically degenerate cluster.
\end{itemize}

Notably, $U[:,26]$ combines audit-related vocabulary (\textit{auditor}, \textit{Auditor}) with identity and authority terms (\textit{evangel}, \textit{imperson}, \textit{Avatar}, \textit{vamp}, \textit{sabot}), suggesting a shared latent direction encoding concepts of \textit{authority, legitimacy, and their subversion}.

\subsection{Gemma-2-2B}
\label{sec:gemma}

The highest-VCS eigenvector ($U[:,4]$, VCS$=0.88$) is dominated by tokens in historical English orthography, specifically the long-s convention used in typography prior to the nineteenth century (\textit{it\longs elf}, \textit{Re\longs}, \textit{Di\longs}, \textit{Hou\longs e}, \textit{An\longs}, \textit{Monfieur}, \textit{my\longs elf}).\footnote{These tokens are reproduced as they appear in Gemma's tokenizer vocabulary. The long-s character shown here (\longs{}) approximates Unicode U+017F (Latin small letter long s), which is present verbatim in the original tokens. The prevalence of this character in the dominant eigenvector is itself informative: it implies that Gemma's pretraining corpus contains a substantial volume of pre-nineteenth-century printed text in which U+017F was used systematically, sufficient to leave a dominant geometric trace in the \texttt{lm\_head} weight matrix.} This is a striking finding: the most geometrically coherent vocabulary subspace in Gemma encodes an archaic writing system no longer in active use. The leading singular vector $U[:,0]$ also encodes historical English orthography with overlapping token sets (VCS$=0.88$).

This observation implies that Google's pretraining corpus for Gemma includes a substantial volume of digitised historical documents, and that this corpus exerts a dominant influence on the directional structure of the \texttt{lm\_head}. Further coherent subspaces include:

\begin{itemize}
    \item $U[:,2]$ (VCS$=0.61$): Contemporary French adjectives in the plural form (\textit{sauvages}, \textit{actuels}, \textit{compl\`{e}tes}, \textit{religieuses}, \textit{automatiques}), indicating a dedicated Romance-language subspace.
    \item $U[:,10]$ (VCS$=0.54$): Historical French vocabulary including past participles and legal register terms (\textit{sup\'{e}rieurs}, \textit{bless\'{e}s}, \textit{ferm\'{e}s}, \textit{refus\'{e}}).
    \item $U[:,14]$ (VCS$=0.28$): Epistemic adverbs encoding evidential stance (\textit{Ironically}, \textit{obviously}, \textit{needlessly}, \textit{unnecessarily}, \textit{conceivably}, \textit{inevitably}). This subspace captures a coherent pragmatic category with direct relevance to logical consistency analysis.
    \item $U[:,7]$ (VCS$=0.33$): Multilingual proper nouns and place names across Hebrew, Ukrainian, Spanish, and Greek.
\end{itemize}

Critically, no eigenvector of Gemma-2-2B was found to contain vocabulary requiring ethical restriction in this paper.

\subsection{Qwen2.5-1.5B}
\label{sec:qwen}

The highest-VCS eigenvectors of Qwen2.5-1.5B ($U[:,7]$, VCS$=0.64$; $U[:,3]$, VCS$=0.66$; $U[:,12]$, VCS$=0.66$; $U[:,13]$, VCS$=0.62$) are dominated by Hangul (Korean script) characters. These clusters exhibit high geometric coherence at the script level but low semantic coherence: the tokens share orthographic properties rather than meaning. This pattern differs qualitatively from the high-VCS clusters of GPT-OSS-120B (syntactic function) and Gemma-2-2B (historical register), suggesting that Qwen's training corpus contains a large volume of Korean text whose embedding geometry was not integrated with semantic structure during training.

The majority of remaining eigenvectors exhibit low VCS (below 0.25), with token clusters spanning unrelated languages, domains, and registers within a single singular direction. This geometric dispersion implies that small perturbations of the hidden state along these directions may activate semantically unrelated tokens, reducing output predictability and auditability.

Most significantly, several eigenvectors of Qwen2.5-1.5B contain vocabulary that the authors have determined to be ethically inappropriate for direct publication in this paper. A detailed discussion of this finding and its implications is provided in Section~\ref{sec:safety}.

\subsection{Singular Value Distribution and Vocabulary Selection Strategy}

\begin{figure}[h]
\centering
\includegraphics[width=\linewidth]{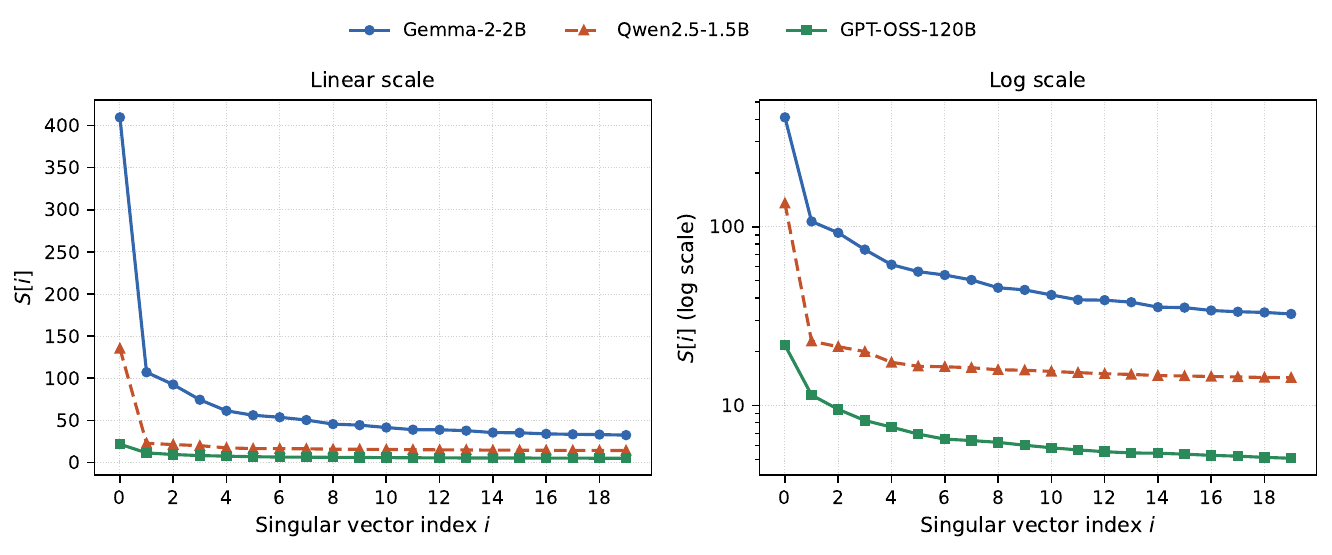}
\caption{Singular value decay curves for the top 20 singular vectors of each model (linear scale, left; log scale, right). Gemma exhibits power-law decay; Qwen shows a sharp cliff after $S[0]$; GPT decays gradually.}
\label{fig:sv_dist}
\end{figure}

Figure~\ref{fig:sv_dist} shows the top-20 singular values for each model. The distributions differ qualitatively across all three models, revealing distinct vocabulary selection strategies encoded in the \texttt{lm\_head} weight matrix.

\begin{table}[h]
\centering
\caption{Singular value statistics for the top 20 eigenvectors.}
\label{tab:sv_stats}
\begin{tabular}{lrrrr}
\toprule
Model & $S[0]$ & $S[1]$ & $S[0]/S[1]$ & Decay pattern \\
\midrule
Gemma-2-2B   & 409.66 & 107.16 & 3.82 & Stepwise clustering \\
Qwen2.5-1.5B & 135.43 &  22.88 & 5.92 & Cliff then plateau \\
GPT-OSS-120B &  21.76 &  11.44 & 1.90 & Gentle slope \\
\bottomrule
\end{tabular}
\end{table}

\paragraph{Gemma: stepwise clustering and high controllability.}
Viewed on a logarithmic scale, Gemma's singular value decay does not follow a power law but instead exhibits a stepwise clustering structure: $S[0] = 409.66$ is strongly dominant, followed by a first cluster ($S[1]$--$S[3]$: 107--74), a second cluster ($S[4]$--$S[9]$: 61--44), and a third cluster ($S[10]$--$S[19]$: 41--32). Whether intentional or emergent from training data composition, this structure has a critical practical implication: vocabulary selection can be controlled by manipulating a small number of dominant singular directions. Because the top singular vectors account for a disproportionate share of spectral energy, targeted interventions on a few eigenvectors suffice to steer the output vocabulary. This architectural property likely contributes to the high VCS values observed for Gemma (Section~\ref{sec:vcs}) and may explain why AOP achieves its highest accuracy on Gemma~\citep{aop2025}: the stepwise structure stabilises $\mathbb{F}_2$ projections by providing a small number of geometrically coherent and dominant directions.

\paragraph{Speculation on design intent (unverified).}
We note, without empirical support, that the stepwise clustering structure described above is unlikely to arise from training data composition alone. One possible---though entirely unverified---interpretation is that Google's pretraining pipeline incorporates objectives or regularisation terms that encourage spectral concentration, thereby building controllability into the \texttt{lm\_head} geometry prior to post-training alignment. The near-identical \texttt{lm\_head} structure between Gemma's base and instruct variants (Section~\ref{sec:base_instruct}) is consistent with this interpretation, as it suggests that the controllable structure is established during pretraining rather than introduced by RLHF. We emphasise that this is purely speculative; verifying or refuting this hypothesis would require access to Google's training methodology, which is not publicly available. We include this speculation solely to motivate future investigation.

\paragraph{Qwen: two-tier structure.}
Qwen shows a sharp cliff after $S[0]$: the ratio $S[0]/S[1] = 5.92$ is the largest among the three models, and $S[1]$ through $S[19]$ form a near-plateau (values between 22.88 and 14.31). This two-tier structure implies that the first singular vector acts as a structural foundation, while the remaining directions contribute roughly equally---a pattern consistent with broad, less curated training data in which no single semantic domain dominates beyond the first. The low VCS values observed for most of Qwen's eigenvectors (Section~\ref{sec:vcs}) are consistent with this interpretation: the plateau directions encode geometrically dispersed vocabulary clusters.

\paragraph{GPT: graduated hierarchy.}
GPT-OSS-120B displays a gentle, monotone decay from $S[0] = 21.76$ to $S[19] = 5.07$, with $S[0]/S[1] = 1.90$---the smallest ratio among the three models. No single direction dominates; instead, each singular vector carries a distinct and progressively smaller share of spectral energy. This graduated structure corresponds to the functionally differentiated subspaces identified in Section~\ref{sec:gpt}: syntactic structure, attribute taxonomy, formal specification vocabulary, software engineering vocabulary, and others each occupy a distinct and well-separated eigenvector. The result is the most \emph{democratically distributed} vocabulary selection architecture of the three models examined.

\paragraph{Summary.}
The singular value distributions expose three distinct design philosophies: Gemma prioritises depth through a small number of dominant, coherent directions; Qwen exhibits a sharp two-tier structure reflecting broad coverage with limited curation; and GPT achieves a functionally graduated hierarchy in which many directions each carry distinct semantic content. These differences are not merely quantitative---they reflect fundamentally different approaches to training data composition and vocabulary organisation, and they directly predict the qualitative differences in eigenvector interpretability reported in Sections~\ref{sec:gpt}--\ref{sec:qwen}.

\subsection{Base--Instruct Comparison}

\paragraph{Gemma-2-2B.}
The \texttt{lm\_head} eigenvector structure of Gemma-2-2B is virtually identical between the base and instruction-tuned variants (VCS differences $< 0.003$ across all 30 singular vectors; token clusters unchanged). This indicates that Gemma's post-training alignment procedure does not modify the vocabulary projection layer, and that instruction-following capability is achieved through modifications to other components of the network.

\paragraph{Qwen2.5-1.5B.}
Qwen2.5-1.5B shows substantial restructuring of \texttt{lm\_head} eigenvectors between base and instruct variants. Most notably, several singular vectors whose top tokens in the base model consist entirely of punctuation and bracket characters are remapped in the instruct variant to multilingual lexical clusters spanning Turkish, Korean, and English. This suggests that Qwen's RLHF procedure modifies the vocabulary projection layer directly, reassigning structural symbol subspaces to conversational lexical subspaces. The VCS of affected eigenvectors changes accordingly (e.g.\ $U[:,1]$: base VCS$=0.33$ with bracket tokens $\rightarrow$ instruct VCS$=0.23$ with multilingual tokens).

Critically, eigenvectors containing ethically concerning vocabulary are present in \emph{both} the base and instruct variants of Qwen2.5-1.5B, with comparable VCS values. This demonstrates that post-training alignment does not remove the problematic subspaces: they originate in pretraining and persist through RLHF.

\paragraph{Summary.}
The base--instruct comparison reveals a three-layer structure in \texttt{lm\_head} eigenvectors:
\begin{enumerate}
    \item \textbf{Tokenizer-determined structure}: script-level clusters (e.g.\ Hangul in Qwen) that are invariant to post-training in both models.
    \item \textbf{Pretraining-determined structure}: semantic and ethical content crystallised from training data, persistent through RLHF in all observed cases.
    \item \textbf{RLHF-induced restructuring}: present in Qwen but absent in Gemma, remapping symbol clusters to multilingual lexical clusters without resolving pretraining-level issues.
\end{enumerate}

\section{Safety Implications}
\label{sec:safety}

\subsection{Ethically Concerning Vocabulary in lm\_head Eigenvectors}

In the course of our analysis, we observed that several eigenvectors of Qwen2.5-1.5B contain vocabulary that the authors have determined to be ethically inappropriate for direct publication in this paper. We document this finding here without reproducing the specific tokens, in accordance with responsible disclosure principles.

The concerning vocabulary appears in eigenvectors with low VCS scores (below 0.20), in the form of geometrically dispersed clusters that span unrelated semantic domains. Critically, these clusters are present in \emph{both} the base and instruction-tuned variants of Qwen2.5-1.5B with comparable VCS values, demonstrating that post-training alignment does not remove the problematic subspaces. The same vocabulary clusters are absent from the corresponding analysis of GPT-OSS-120B and Gemma-2-2B.

We emphasise that \texttt{lm\_head} SVD analysis is a static, weight-only procedure: it does not require model inference, prompt engineering, or any form of adversarial attack. The fact that ethically concerning vocabulary is directly surfaced by this analysis---without any attempt to elicit it---is itself a significant safety-relevant observation.

\subsection{A Taxonomy of Root Causes}

The presence of ethically concerning vocabulary in a model's \texttt{lm\_head} eigenvectors implies that such vocabulary was present in the pretraining corpus with sufficient frequency and co-occurrence density to leave a geometric trace in the weight matrix. Beyond this, the underlying cause may vary. We propose the following taxonomy, not to assign responsibility in any specific case, but to structure future investigation and inform governance discussions.

\begin{description}
    \item[Category A: Deliberate inclusion.]
    The training data containing the problematic vocabulary was intentionally selected or retained.

    \item[Category B: Known omission.]
    The presence of problematic content in the training data was known to the developers, but removal was not prioritised.

    \item[Category C: Inadvertent inclusion.]
    The problematic content was included without awareness, due to insufficient data auditing or filtering pipelines.

    \item[Category D: Definitional divergence.]
    The vocabulary in question is not universally recognised as problematic; cultural, linguistic, or jurisdictional differences in what constitutes harmful content may account for its presence.
\end{description}

We make no determination as to which category applies to the observations reported in this paper. The distinction between categories matters considerably from a governance perspective: Category A implies legal liability; Category B implies accountability failures; Category C implies process failures that can be addressed through improved tooling; and Category D implies the need for internationally harmonised standards. Regardless of which category applies in a given case, the outcome for downstream users is the same, and the detection method we propose is applicable in all cases.

We note that the absence of an international regulatory framework governing training data composition means that, at present, there is no mechanism to compel disclosure or remediation in any of these categories. This observation motivates the call to action in Section~\ref{sec:conclusion}.

\subsection{RLHF Does Not Resolve Pretraining-Level Issues}

Our base--instruct comparison (Section~\ref{sec:base_instruct}) provides direct evidence that post-training alignment through RLHF does not remove ethically concerning vocabulary subspaces from the \texttt{lm\_head}. The problematic eigenvectors persist across both variants with comparable structure and VCS values.

This finding has a practical implication for the current practice of AI safety: if safety evaluation is conducted only on instruction-tuned models, pretraining-level issues may go undetected. We recommend that \texttt{lm\_head} SVD analysis be applied to base models as a standard pre-release evaluation step, prior to any post-training procedure.

\subsection{lm\_head SVD as a Safety Auditing Tool}

The analysis presented in this paper requires only the \texttt{lm\_head} weight matrix and the model's tokenizer. It does not require:
\begin{itemize}
    \item model inference or GPU execution beyond weight loading,
    \item labeled evaluation datasets,
    \item adversarial prompts or red-teaming,
    \item access to training data.
\end{itemize}

This simplicity is both a strength and an argument for adoption: the barrier to running this analysis is low enough that it can reasonably be incorporated into any model release pipeline. We propose VCS-based eigenvector analysis as a lightweight complement to existing safety benchmarks, not a replacement.

Existing glitch token detection methods---including GlitchHunter~\citep{li2024glitchhunter}, GlitchProber~\citep{zhang2024glitchprober}, and GlitchMiner~\citep{liu2024glitchminer}---all require model inference and produce candidate lists that still require human judgement to interpret. As noted by a practitioner in the Japanese LLM development community\footnote{See \url{https://x.com/imos/status/1966886238448959855}}, manually reviewing the full vocabulary of a model (on the order of 100,000 tokens) is feasible in roughly half a day at a rate of ten tokens per second---and is currently the most reliable approach for ensuring vocabulary quality prior to release. \texttt{lm\_head} SVD analysis can complement this human review process: by identifying eigenvectors with low VCS or ethically concerning token clusters, it can generate a prioritised list of tokens warranting closer human inspection, potentially concentrating the manual effort where it is most needed rather than eliminating it.

\subsection{Relationship to Algebraic Ontology Projection}

The Algebraic Ontology Projection (AOP) framework~\citep{aop2025} operates on hidden state projections rather than on vocabulary distributions, and is therefore structurally independent of the \texttt{lm\_head} eigenvector pathologies documented in this paper. This independence has two implications.

First, AOP can serve as an objective observer of model behaviour even in models whose \texttt{lm\_head} structure is compromised, potentially detecting logical inconsistencies that originate from incoherent vocabulary subspaces.

Second, the algebraic constraints imposed by AOP on prompt design suggest a potential pathway toward mitigating---though not eliminating---the influence of low-VCS vocabulary subspaces on model output. A full investigation of this possibility is left for future work.

\section{Glitch Token Detection via Weighted Projection Score}
\label{sec:glitch}

\subsection{Motivation}

Glitch tokens are vocabulary entries that cause LLMs to produce erratic or incoherent outputs when presented as input~\citep{rumbelow2023solidgoldmagikarp,li2024glitchhunter}. Their root cause is a mismatch between the tokenizer training corpus and the LLM pretraining corpus: tokens that appear in the vocabulary but rarely in the pretraining data acquire poorly defined embeddings, rendering them geometrically isolated in the \texttt{lm\_head} weight space.

We note that glitch tokens are distinct from the ethically concerning vocabulary discussed in Section~\ref{sec:safety}: the latter originates from \emph{over-representation} in the pretraining corpus, whereas glitch tokens arise from \emph{under-representation}. Nevertheless, the same SVD framework introduced in this paper suggests a lightweight static approach to glitch token detection that does not require model inference.

\subsection{Weighted Projection Score}

For each token $v$ in the vocabulary, we define its \emph{Weighted Projection Score} (WPS) as the sum of its absolute contributions across all singular directions, weighted by the corresponding singular values:

\begin{equation}
    \text{WPS}(v) = \sum_{k=1}^{d} S[k] \cdot |U[v, k]|
\end{equation}

Intuitively, $\text{WPS}(v)$ measures how strongly token $v$ participates in the dominant directions of the \texttt{lm\_head} weight space. Tokens with high WPS are readily activated by the model's hidden states; tokens with low WPS are geometrically marginalised---their \texttt{lm\_head} rows point in directions that carry little singular value energy---and are therefore candidates for glitch token status.

We propose a threshold of $\mu - 2\sigma$ (where $\mu$ and $\sigma$ are the mean and standard deviation of WPS across the full vocabulary) as a practical cutoff, identifying approximately the bottom 2--4\% of tokens as candidates.

The implementation requires no inference and adds only a few lines to the existing SVD analysis:

\begin{lstlisting}[language=Python]
# U: (V, d), S: (d,)
weighted_u = U.abs() * S.unsqueeze(0)   # (V, d)
wps = weighted_u.sum(dim=1)             # (V,)

mean, std = wps.mean(), wps.std()
threshold = mean - 2.0 * std
glitch_candidates = (wps < threshold).nonzero(as_tuple=True)[0]
\end{lstlisting}

\subsection{Preliminary Results}

We applied this procedure to Gemma-2-2B and Qwen2.5-1.5B (instruct variants) and report summary statistics in Table~\ref{tab:glitch_stats}.

\begin{table}[h]
\centering
\caption{Glitch token candidate statistics ($\mu - 2\sigma$ threshold).}
\label{tab:glitch_stats}
\begin{tabular}{lrrrr}
\toprule
Model & $\mu$ & $\sigma$ & Threshold & Candidates (\%) \\
\midrule
GPT-OSS-120B  &  9.95 & 2.94 &  4.07 & 9,819 (4.88\%) \\
Gemma-2-2B    & 58.58 & 7.32 & 43.93 & 5,543 (2.17\%) \\
Qwen2.5-1.5B  & 29.23 & 5.30 & 18.62 & 5,885 (3.87\%) \\
\bottomrule
\end{tabular}
\end{table}

The qualitative character of the candidates differs markedly across models.

For GPT-OSS-120B, the lowest-scoring tokens (WPS $\approx 0.63$) are token IDs in the range 200000--201087 that decode as empty strings---consecutive reserved entries at the tail of the vocabulary. More diagnostically, tokens with WPS $\approx 0.87$ include a dense cluster of Chinese-language gambling and lottery spam strings (tokens containing phrases meaning ``daily lottery'', ``fast three lottery'', and ``lottery god'' and their variants), full-width Japanese katakana, and---notably---the token \glitchtoken{} (ID 137606, WPS = 0.8807), a glitch token widely known in the CJK language community\footnote{See \url{https://note.com/xcloche/n/n55938e706986} for a detailed account in Japanese.}. Its detection by WPS without any inference confirms that the approach captures known problematic tokens.

For Gemma-2-2B, the lowest-scoring tokens are predominantly single ASCII characters, control characters (U+0001--U+001F), and instances of the Unicode replacement character U+FFFD assigned to multiple token IDs---suggesting redundant tokenizer entries rather than glitch tokens in the classical sense.

For Qwen2.5-1.5B, a large cohort of tokens score at virtually the same minimum value (WPS $\approx 3.01$), comprising Thai-script tokens (e.g., tokens meaning ``that'', ``have'', and ``all'' in Thai), Arabic diacritics, Hebrew sequences, and isolated CJK characters. Additionally, token IDs in the high-index range (151829--151860) decode as empty or non-displayable strings---consecutive reserved entries at the tail of the vocabulary that are entirely absent from pretraining data. Further examples include IDE-derived compound tokens such as \texttt{ForCanBeConvertedToF} and full-width numeric forms such as full-width \texttt{20}. In total, approximately 6,000 tokens (3.87\% of the 151,936-entry vocabulary) fall below the threshold, suggesting that a non-trivial fraction of Qwen's vocabulary is effectively unused by the model.

This pattern---many tokens collapsed to the same near-zero score---is consistent with the embedding-space clustering reported by GlitchHunter~\citep{li2024glitchhunter}: tokens registered in the tokenizer but absent from pretraining data are assigned degenerate \texttt{lm\_head} rows that all point in the same ``garbage'' direction.

These observations suggest that WPS provides a complementary static signal to existing inference-based detection methods. Unlike GlitchHunter, which constructs a token embedding graph and applies the Leiden clustering algorithm~\citep{traag2019leiden}, WPS requires no inference and produces a ranked list of candidates in a single pass over the \texttt{lm\_head} weight matrix.

\subsection{Limitations and Future Work}

The WPS approach has two important limitations. First, it detects tokens that are \emph{geometrically marginalised} in the \texttt{lm\_head} space, which correlates with but does not perfectly identify glitch tokens: some low-WPS tokens may be legitimately rare vocabulary items rather than pathological entries. Second, the threshold $\mu - 2\sigma$ is a heuristic; a systematic evaluation against the labelled glitch token datasets of~\citet{li2024glitchhunter} and~\citet{zhang2024glitchprober} is needed to calibrate precision and recall.

We also note that WPS is entirely independent of token content: it cannot distinguish between a harmless rare token and an ethically problematic one. The two problems---geometric marginalisation and ethical content---require separate analysis pipelines, as discussed in Section~\ref{sec:safety}.

A full empirical evaluation of WPS as a glitch token detector, including comparison with GlitchHunter, GlitchProber~\citep{zhang2024glitchprober}, and GlitchMiner~\citep{liu2024glitchminer}, is left for future work.

\section{Related Work}
\label{sec:related}

\subsection{Glitch Tokens}

Glitch tokens are vocabulary entries that cause LLMs to produce erratic, incoherent, or policy-violating outputs when encountered as input. The phenomenon was first systematically documented by Rumbelow and Watkins~\citep{rumbelow2023solidgoldmagikarp}, who identified a set of anomalous tokens in GPT-2 and GPT-3 that clustered near the centroid of the embedding space and could not be correctly repeated by the model. The underlying mechanism is a mismatch between the tokenizer training corpus and the LLM pretraining corpus: tokens that appear frequently enough to be included in the vocabulary but rarely enough in the LLM training data to acquire meaningful embeddings become geometrically isolated, resulting in unpredictable model behaviour~\citep{rumbelow2023solidgoldmagikarp}.

Li et al.~\citep{li2024glitchhunter} conducted the first comprehensive empirical study of glitch tokens across seven LLMs and 182,517 tokens, proposing GlitchHunter: a detection method based on constructing a token embedding graph and applying the Leiden clustering algorithm~\citep{traag2019leiden} to identify dense anomalous clusters. Zhang et al.~\citep{zhang2024glitchprober} proposed GlitchProber, which instead analyses internal activations (hidden states and attention outputs) using PCA and SVM classifiers, and additionally provides a mitigation mechanism that masks offending neurons at inference time. More recently, Liu et al.~\citep{liu2024glitchminer} proposed GlitchMiner, a gradient-based discrete optimisation framework that detects glitch tokens by maximising predictive entropy, overcoming the dependence on predefined embedding patterns shared by earlier methods.

Our work relates to this literature in two ways. First, low-VCS eigenvectors in our analysis may overlap with the embedding-space anomalies exploited by GlitchHunter, suggesting that \texttt{lm\_head} SVD analysis could serve as a static, inference-free complement to existing detection methods. Second, and importantly, the phenomena are \emph{not} identical: glitch tokens arise from \emph{under-representation} in the pretraining corpus (too few occurrences to form stable embeddings), whereas the ethically concerning vocabulary clusters we observe in Qwen2.5-1.5B appear in \emph{upper} eigenvectors with substantial singular values, implying \emph{over-representation}---sufficient frequency and co-occurrence density to leave a dominant geometric trace in the \texttt{lm\_head} weight matrix. These are complementary failure modes with different root causes and different implications for remediation.

\subsection{Interpreting the lm\_head and Vocabulary Projection}

The Logit Lens~\citep{nostalgebraist2020logitlens} projects intermediate hidden states through the \texttt{lm\_head} to visualise how token predictions evolve across layers, providing a tool for mechanistic interpretability. The Tuned Lens~\citep{belrose2023tuned} extends this by learning an affine transformation per layer to improve the quality of intermediate predictions. The Backward Lens~\citep{backwardlens2024} inverts the direction of analysis, projecting gradients rather than activations into vocabulary space.

Our work is complementary but distinct: rather than projecting hidden states \emph{through} the \texttt{lm\_head}, we analyse the \texttt{lm\_head} weight matrix \emph{itself} via SVD. This requires no input data and no model inference, and reveals the static geometric structure of vocabulary selection independent of any particular hidden state.

The softmax bottleneck~\citep{yang2018breaking} identifies a fundamental limitation of the standard LM head architecture: when the hidden dimension $d$ is smaller than the vocabulary size $V$, the expressiveness of the output distribution is constrained. Our singular value analysis provides a complementary view of this bottleneck: the effective rank of $W$, reflected in the decay profile of singular values, determines how many independent vocabulary selection directions are available to the model.

\subsection{Training Data Quality and Safety}

The quality and composition of pretraining data have been shown to have lasting effects on model behaviour that are difficult to reverse through post-training alignment~\citep{longpre2023pretrainer}. Our base--instruct comparison provides direct empirical support for this claim at the level of \texttt{lm\_head} geometry: problematic vocabulary subspaces present in base models persist through RLHF.

This persistence is consistent with recent findings on the sparsity of alignment updates. Mukherjee et al.~\citep{mukherjee2025subnetworks} show that even with full-parameter finetuning, reinforcement learning updates only 5--30\% of parameters (sparse subnetworks), with the remainder essentially unchanged. By matrix perturbation theory (Davis--Kahan theorem), adding a sparse, small-norm perturbation to a large pretraining matrix is unlikely to rotate its dominant singular vectors significantly---providing a theoretical account of why \texttt{lm\_head} eigenvector structure is largely preserved after RLHF.

To our knowledge, no prior work has directly tracked changes in \texttt{lm\_head} singular vectors between base and instruction-tuned variants at the token level. Our base--instruct comparison provides the first such direct evidence, connecting the macroscopic sparsity statistics of~\citet{mukherjee2025subnetworks} to microscopic geometric changes in the vocabulary projection layer.

\subsection{Tokenizer Optimisation and Vocabulary Efficiency}

The design of LLM tokenizers has received increasing attention as a source of both performance and efficiency improvements. Standard subword algorithms such as Byte Pair Encoding (BPE)~\citep{sennrich2016bpe} construct vocabularies based on corpus frequency statistics, without regard to how the resulting vocabulary will interact with the model's weight matrices during training.

\subsection{Tokenizer Optimisation and Vocabulary Efficiency}

Tokenizer design has received increasing attention as a source of LLM efficiency improvements. Standard BPE~\citep{sennrich2016bpe} constructs vocabularies from corpus frequency statistics alone. Recent work has explored several directions beyond this baseline.

Frequency and compression-based approaches include BoundlessBPE~\citep{boundlessbpe2025}, which relaxes pre-tokenization boundary constraints to allow cross-word merges, and LiteToken~\citep{litetoken2026}, which identifies and removes ``intermediate merge residues''---tokens that are frequent during BPE training but rarely appear in final tokenized output, affecting approximately 10\% of tokens in major tokenizers.

Model-feedback-based approaches use LLM inference signals to guide vocabulary construction. Zheng et al.~\citep{zheng2024adat} (NeurIPS 2024) propose an adaptive tokenizer that monitors changes in model perplexity during training to iteratively refine the vocabulary, selecting tokens that are closely aligned with the model's evolving dynamics. Retrofitting LLMs with Dynamic Tokenization~\citep{dyntoken2025} (ACL 2025) enables flexible post-training tokenization by adaptively choosing token granularity during inference.

Domain and language adaptation approaches~\citep{tokalign2024,vocabcustom2024} extend or replace tokenizer vocabularies to improve compression rates for low-resource languages and specialised domains.

Across all of these approaches, the feedback signal is derived from either the training corpus (frequency, compression ratio) or model inference (loss, latency). \emph{None uses the static geometric structure of the \texttt{lm\_head} weight matrix as a vocabulary quality criterion.} Our proposed SVD-guided tokenizer optimisation direction fills this gap: the VCS metric provides a purely static, inference-free signal that reflects the geometric stability of vocabulary selection in the output projection layer. To our knowledge, this connection between \texttt{lm\_head} eigenvector analysis and tokenizer design has not been previously proposed.

Vocabulary adaptation for domain transfer and low-resource languages has also been studied extensively~\citep{tokalign2024,vocabcustom2024}, primarily motivated by compression efficiency (reducing sequence length) rather than vocabulary quality or geometric stability. Our motivation is orthogonal: we aim to identify tokens that are geometrically stable in the \texttt{lm\_head} eigenspace, which we conjecture corresponds to tokens that the model can select reliably and consistently.

To our knowledge, no prior work has proposed using \texttt{lm\_head} SVD analysis as a criterion for tokenizer construction or vocabulary pruning. The potential connection between high-VCS token selection and parameter efficiency---smaller vocabulary sizes with higher per-token representational quality---remains an open empirical question that we leave to future work.

The Algebraic Ontology Projection (AOP) framework~\citep{aop2025} projects LLM hidden states onto the binary field $\mathbb{F}_2$ and measures the algebraic consistency of \textit{is-a}/\textit{has-a} ontological relations via the Semantic Crystallisation (SC) metric. AOP demonstrated zero-shot inclusion accuracy of up to 93.33\% across multiple model families without fine-tuning, and identified Late-layer Collapse as a failure mode in which logical consistency degrades in intermediate layers. The present paper identifies a structural property of the \texttt{lm\_head}---the VCS distribution across eigenvectors---that is complementary to the hidden-state analysis of AOP and may help explain why certain models are more amenable to AOP-based prompt optimisation than others.

\section{Conclusion}
\label{sec:conclusion}

We have shown that singular value decomposition of the \texttt{lm\_head} weight matrix of a transformer-based LLM reveals interpretable semantic subspaces directly, without model inference, prompt engineering, or access to training data. The left singular vectors partition the vocabulary into geometrically coherent clusters whose character reflects the composition and curation philosophy of the pretraining corpus. We introduced the Vocabulary Cluster Score (VCS) to quantify this coherence, and demonstrated its utility across three models of different provenance and scale.

Our principal findings are as follows.

\begin{enumerate}

\item \textbf{lm\_head eigenvectors encode interpretable semantic subspaces.}
Each singular vector concentrates vocabulary from a distinct semantic, linguistic, or functional domain. The nature of these domains differs systematically across models: GPT-OSS-120B exhibits functionally differentiated subspaces (syntactic structure, formal specification vocabulary, software engineering terminology); Gemma-2-2B exhibits historically and linguistically stratified subspaces dominated by pre-nineteenth-century printed English; and Qwen2.5-1.5B exhibits script-level clustering alongside geometrically dispersed, low-VCS subspaces.

\item \textbf{Singular value distributions encode vocabulary selection strategy.}
The decay profile of singular values reflects the degree to which vocabulary selection is concentrated in a small number of dominant directions. Gemma exhibits stepwise clustering that confers high controllability; Qwen exhibits a sharp two-tier structure; and GPT exhibits a graduated hierarchy in which many directions carry distinct semantic content.

\item \textbf{Pretraining-level content persists through RLHF.}
Base--instruct comparison reveals that ethically concerning vocabulary subspaces, where present, are not removed by post-training alignment. RLHF in Qwen restructures some eigenvectors (remapping symbol clusters to multilingual lexical clusters) but does not resolve the underlying pretraining-level issues. In Gemma, RLHF leaves the \texttt{lm\_head} structure virtually unchanged.

\item \textbf{lm\_head SVD analysis is a viable safety auditing tool.}
The analysis requires only the weight matrix and tokenizer, runs in minutes, and surfaces ethically concerning vocabulary without any form of adversarial prompting. We propose it as a lightweight complement to existing safety evaluation pipelines, to be applied to base models prior to post-training.

\end{enumerate}

\paragraph{Limitations.}
VCS measures geometric coherence in the \texttt{lm\_head} weight space, which encompasses semantic, script-level, and functional similarity without distinguishing between them. Qualitative inspection of decoded tokens is therefore necessary to interpret VCS values. Our analysis covers three models; broader coverage across architectures, scales, and training paradigms is needed to generalise the findings. The relationship between \texttt{lm\_head} eigenvector structure and downstream model behaviour---including the connection to AOP and logical consistency---remains an empirical question for future work.

Regarding the SVD-guided tokenizer optimisation direction described below: the proposed Best Token Extraction approach selects tokens based on geometric stability alone, without regard to content. It will therefore select high-VCS tokens regardless of whether they are benign or harmful; it cannot be used as a safety auditing tool. Furthermore, the method operates at the level of token selection only: it cannot perform vocabulary restructuring operations such as merging tokens into more appropriate granularities or splitting coarse tokens. These limitations constrain its applicability to tokenizer quality improvement rather than safety assurance.

\paragraph{Future work.}
The VCS metric introduced in this paper suggests a principled approach to tokenizer design. Eigenvectors with high VCS identify vocabulary clusters that are geometrically stable in the \texttt{lm\_head} weight space---tokens that the model can select reliably and consistently. Conversely, tokens that appear only in low-VCS, low-eigenvalue eigenvectors may be candidates for removal or replacement.

We term this direction \textit{SVD-guided tokenizer optimisation} and sketch its iterative form:
\begin{enumerate}
    \item Train an initial LLM with a standard tokenizer.
    \item Apply \texttt{lm\_head} SVD analysis to compute per-token VCS contributions.
    \item Reconstruct the tokenizer, prioritising high-VCS tokens and removing low-VCS candidates.
    \item Retrain the LLM with the updated tokenizer.
    \item Repeat until convergence.
\end{enumerate}
If effective, this procedure would build vocabulary selection efficiency into the model from the ground up, potentially yielding a singular value distribution resembling Gemma's stepwise clustering structure with fewer training iterations. We leave empirical validation of this hypothesis to future work.

A second direction concerns the relationship between \texttt{lm\_head} eigenvector structure and glitch tokens~\citep{rumbelow2023solidgoldmagikarp,li2024glitchhunter}. Tokens that cluster in low-VCS eigenvectors may overlap with the anomalous tokens identified by embedding-space methods such as GlitchHunter~\citep{li2024glitchhunter}. Unlike GlitchHunter, which requires inference, \texttt{lm\_head} SVD analysis is entirely static; whether it can serve as a computationally cheaper proxy for glitch token detection remains to be verified.

A third direction is the integration of \texttt{lm\_head} SVD analysis with the Algebraic Ontology Projection framework~\citep{aop2025}, both for prompt optimisation and for the detection of logical inconsistencies arising from low-VCS vocabulary subspaces.

\paragraph{Call to action.}
We call upon the research community to adopt \texttt{lm\_head} SVD analysis as a standard component of model evaluation prior to public release. The simplicity of the method---only five lines of PyTorch---means that the barrier to adoption is low. We further call upon standards bodies and regulatory authorities to consider \texttt{lm\_head} eigenvector analysis as a basis for training data auditing requirements, and upon the broader AI community to develop internationally harmonised standards for what constitutes ethically unacceptable content in pretraining corpora. The taxonomy of root causes proposed in Section~\ref{sec:safety} is intended as a starting point for that discussion.

\section*{Acknowledgements}
The authors thank Taichi Kawabata for insightful discussions on  practical challenges in LLM agent deployment that motivated this work.

\bibliography{aop}
\bibliographystyle{iclr2026_conference}

\appendix
\appendix

\clearpage

\section{Detailed Eigenvector Analysis: GPT-OSS-120B}
\label{app:gpt}

The following table presents the top-15 tokens and VCS for the first 20 singular vectors of GPT-OSS-120B (instruct variant). Token strings are Python \texttt{repr()} output; the integer suffix denotes the ratio of each token's score to the maximum score in that vector (tokens below 10\% are omitted). Singular values $S[i]$ are shown in the \textit{S[i]} column.

\begin{figure}[h]
\centering
\includegraphics[width=\linewidth]{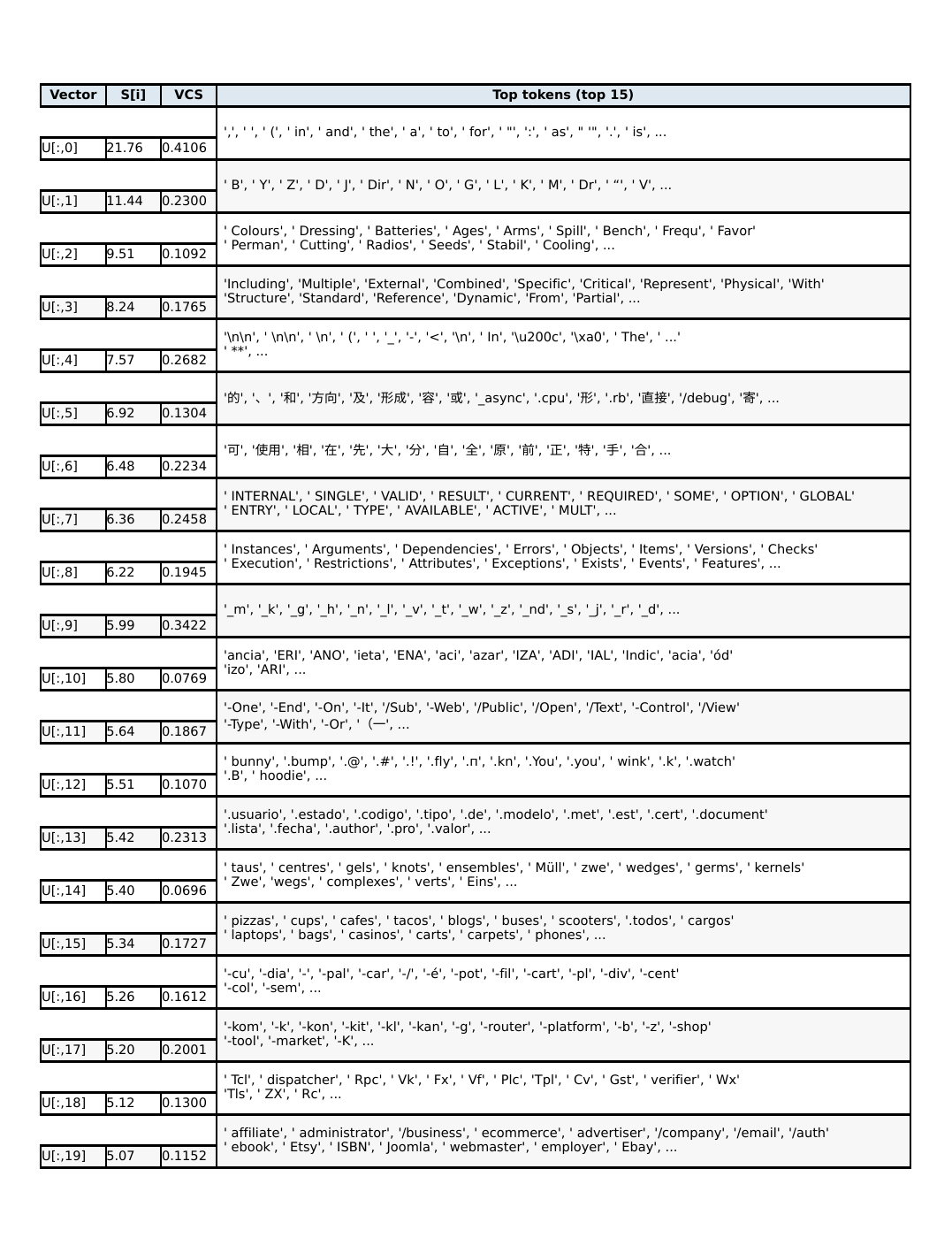}
\caption{GPT-OSS-120B (instruct): top-15 tokens per singular vector,
$U[:,0]$--$U[:,19]$.}
\label{fig:app_gpt}
\end{figure}

\clearpage

\section{Detailed Eigenvector Analysis: Gemma-2-2B}
\label{app:gemma}

The following table presents the top-15 tokens and VCS for the first 20 singular vectors of Gemma-2-2B (instruct variant). Tokens containing Unicode U+017F (Latin small letter long s) are reproduced verbatim as they appear in the model vocabulary; see the footnote in Section~\ref{sec:results} for details.

\begin{figure}[h]
\centering
\includegraphics[width=\linewidth]{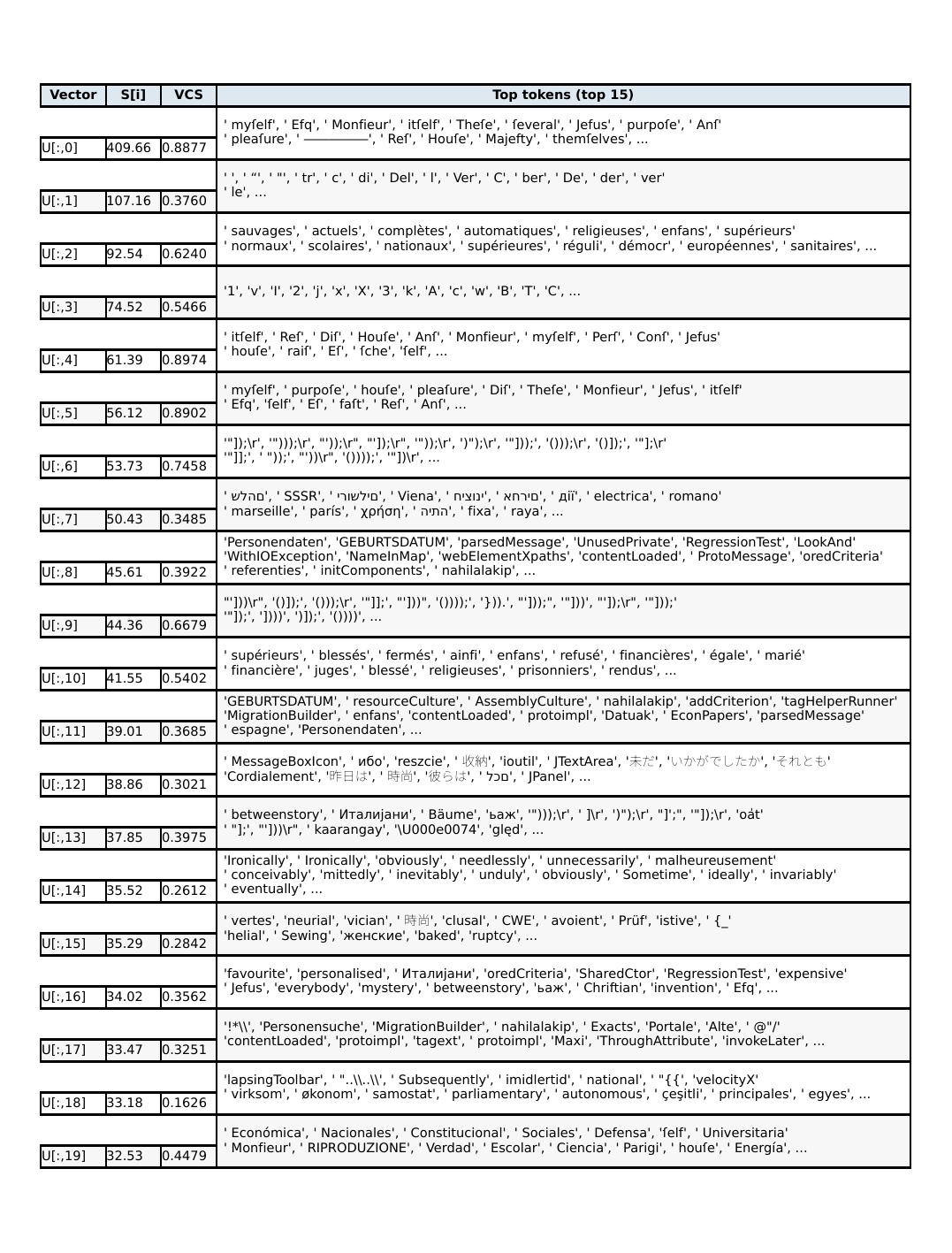}
\caption{Gemma-2-2B (instruct): top-15 tokens per singular vector,
$U[:,0]$--$U[:,19]$.}
\label{fig:app_gemma}
\end{figure}

\clearpage

\section{Note on Qwen2.5-1.5B Appendix}
\label{app:qwen}

Detailed token-level data for Qwen2.5-1.5B are not reproduced in this appendix. As noted in Section~\ref{sec:safety}, several eigenvectors of this model contain vocabulary that the author have determined to be ethically inappropriate for direct publication. In accordance with responsible disclosure principles, we provide only the VCS values and qualitative descriptions in the main text. Researchers requiring the full token lists for verification or safety research purposes may contact the author directly.

\end{document}